\documentclass[conference]{IEEEtran}

\usepackage{booktabs}
\usepackage{graphicx}
\usepackage{amsmath}
\usepackage{amssymb}
\usepackage{multirow}
\usepackage{url}
\usepackage{morefloats}   
\usepackage[hidelinks]{hyperref}
\graphicspath{{figures/}}

\begin{document}

\title{Pre-Inference Routing for Cost-Efficient Document Field Extraction}

\author{
\IEEEauthorblockN{Sreerekha Rajendran}
\IEEEauthorblockA{Independent Researcher \\
Salt Lake City, Utah, United States \\
sreerekha.raju@gmail.com}
}

\maketitle

\begin{abstract}
Most document-extraction systems use a single model for all documents. This is
simple but can be costly for easy cases and less effective for difficult ones. We
examine whether we can predict a document's difficulty \emph{before} extraction
using inexpensive, document-based signals, and use this to choose between a cheaper
and a stronger extractor. We find that routing only helps if two conditions hold:
the cheaper model fails often enough to make routing worthwhile, and those failures
can be predicted from visible features such as image quality and layout. We turn
these into a practical test and apply it to five genres. When both
conditions are met, the calibrated router reduces cost by \textbf{31--33\% on
receipts} and \textbf{77\% on degraded ad-buy forms} while keeping quality within
$0.02$ F1 of always choosing the large model. Routing does not help if either
condition is missing, as with clean digital invoices or nutrition labels that are
already easy to read. A small labeled pilot can predict whether routing will work,
and in the two cases where we ran it first, the prediction was correct. A simple
bag-of-words router works about as well as engineered features, showing that the
main limit is the genre, not the router design; we use interpretable
features to help explain which genres can be routed. The router must be retrained
for each dataset and does not transfer across datasets, even within the same genre.
These results hold for two model pairs with cost differences of $5\times$ and
$3\times$.
\end{abstract}


\section{Introduction}
Organizations handle many kinds of documents, such as invoices, receipts, forms,
and contracts, and often use a single extraction model for all of them. This can
make simple documents more expensive to process, and it limits the benefit of
stronger models to only the hardest cases. We examine whether extraction
difficulty can be predicted \emph{in advance} from low-cost, document-based
features, and used to reduce cost while maintaining accuracy. We find that
pre-inference routing only works when there is a clear difference in difficulty
between documents and when this difference is evident in observable features.

We turn this into a test that can be run on a small labeled sample before rolling
out at scale, and we try it on five genres, including degraded ad-buy
forms, where routing cut cost by $77\%$. In structured extraction the schema stays
fixed while document complexity varies, so routing is based on the document's
characteristics rather than the extraction request. Unlike cascades or deferral
systems, which may run more than one model on a single document, our router
selects one model per document from its features alone.

\noindent\textbf{Contributions.}
\begin{itemize}
  \item A two-part test, which assesses how many documents are difficult and whether features can predict this, lets us decide from a small labeled sample whether pre-inference routing will save cost before full deployment. We ran it on five genres, and in both cases where we ran a pilot first, the prediction was correct.
  \item Strong results on a non-receipt genre: for degraded ad-buy forms the router reduced cost by $77\%$ while keeping quality within $0.02$ F1 of always using the large model (held-out AUC $0.92$); savings on receipts were $31$--$33\%$. Two other genres did not benefit, each missing one of the two test conditions.
  \item The main limit is the genre, not the router. A simple bag-of-words text router works about as well as our interpretable pre-inference features; we focus on the interpretable features because they help explain which genres are good candidates for routing.
  \item The approach is deployment-specific: the router does not generalize across datasets, even within the same genre (for example, CORD to SROIE AUC is about $0.55$), so it must be retrained for each new deployment on that corpus's own labeled sample. We use a field-type canonical scorer and a leakage-resistant feature set, so our numbers are reported conservatively.
  \item Under full-document re-extraction and the same escalation decision, pre-inference routing is consistently cheaper than a confidence cascade because it avoids the extra cheap-model call on escalated documents. This holds for both model pairs, at $5\times$ and $3\times$ cost differences.
\end{itemize}

\section{Related Work}
\label{sec:related}

\paragraph{LLM routing and model selection.}
A growing body of work routes each query to the cheapest model that can answer it
adequately. CARROT~\cite{carrot} frames this as a minimax-optimal selection
problem; MixLLM~\cite{mixllm} and Route-to-Reason~\cite{rtr} route over mixed
model pools, the latter also selecting a reasoning strategy. BEST-Route~\cite{bestroute}
couples routing with test-time compute, and a family of recent
systems~\cite{adaptiveroute,orchestration,agentcost}
target cost-efficient orchestration across models, strategies, and agents. These
routers estimate difficulty by embedding queries or predicting model performance,
and are tested on text QA, reasoning, or multimodal QA~\cite{mmrbench}; surveys
have recently catalogued this space~\cite{routingsurvey}. However, none estimates
difficulty from \emph{intrinsic, document-level} features computable
\emph{pre-inference}, and there is no routing benchmark for structured document
extraction.

\paragraph{Cost-aware cascades and deferral.}
Cascades invoke a cheap model first and escalate based on its output. FrugalGPT~\cite{frugalgpt}
learns a cascade of LLM calls with a scoring function; RouteLLM~\cite{routellm}
and Hybrid-LLM~\cite{hybridllm} learn binary routers between a weak and a strong
model. Crucially, these decisions are made \emph{after} generation (on output
confidence or a learned score), whereas we estimate difficulty \emph{pre-inference}
from the document itself, avoiding a speculative first generation. Our routing
target, escalate only when the strong model's expected F1 gain exceeds a
threshold $\tau$, is an instance of \emph{learning to defer}~\cite{defer} and
selective prediction~\cite{selective}.

\paragraph{Document AI and key-information extraction.}
Structured extraction from visually-rich documents is dominated by layout-aware
models such as LayoutLMv3~\cite{layoutlmv3} and OCR-free transformers such as
Donut~\cite{donut}. We evaluate on the standard receipt benchmarks
CORD~\cite{cord} and SROIE~\cite{sroie} and the more complex, cross-domain
VRDU~\cite{vrdu}; FUNSD~\cite{funsd}, another common benchmark, is incompatible
with our extraction-F1 protocol (Sec.~\ref{sec:limitations}). Most prior work uses
these datasets to improve \emph{extraction accuracy}. We instead hold the extractor
fixed and ask a different question: which documents justify paying for the stronger
tier?

\paragraph{Document complexity and quality signals.}
Our features build on established quality signals: OCR confidence as a proxy for
input degradation, document image-quality assessment (blur, skew), and
layout-complexity measures. We aggregate thirteen such signals into a
pre-inference difficulty estimate. We are not aware of earlier work that uses this
combination of intrinsic signals for \emph{model routing} rather than for quality
filtering or re-scanning.

\section{Method}

\subsection{Problem formulation}
Let $d$ be a document and $x(d)\in\mathbb{R}^{13}$ its pre-inference feature
vector (Sec.~\ref{sec:features}). Two extractors are available: a cheap tier $s$
and an expensive tier $\ell$, yielding field-level F1 quality
$q_s(d),q_\ell(d)\in[0,1]$ at per-document cost $c_s(d),c_\ell(d)$. The
\emph{difficulty gap} is
\begin{equation}
g(d) = q_\ell(d) - q_s(d),
\end{equation}
and the oracle tier label is $y(d)=\mathbf{1}[\,g(d) > \tau\,]$: a document is
\emph{large-required} only if the expensive tier improves F1 by more than the
oracle threshold $\tau$. A router estimates
$p(d)=\widehat{\Pr}[\,y(d){=}1\mid x(d)\,]$ and routes to the large tier when
$p(d)\ge t$; write $r_t(d)=\mathbf{1}[\,p(d)\ge t\,]$. Over $N$ documents,
\begin{align}
Q(t) &= \tfrac1N\textstyle\sum_d \big[r_t(d)\,q_\ell(d) + (1{-}r_t(d))\,q_s(d)\big],\\
C(t) &= \tfrac1N\textstyle\sum_d \big[r_t(d)\,c_\ell(d) + (1{-}r_t(d))\,c_s(d)\big].
\end{align}
Sweeping $t\in[0,1]$ traces the cost--quality Pareto frontier. The target
operating point is the cost-minimal threshold that preserves quality within a
tolerance $\delta$ of always-large:
\begin{equation}
t^\star = \arg\min_{t}\, C(t)\quad\text{s.t.}\quad Q(t)\ge Q_\ell-\delta,
\label{eq:operating}
\end{equation}
where $Q_\ell$ is the always-large mean quality and $\delta=0.02$ (two F1
points). The reported saving is $1 - C(t^\star)/C_\ell$, with $C_\ell$ the
always-large cost. We report the full frontier and the threshold-free AUC as our
primary, selection-free results; the single point $t^\star$ summarises the
frontier at tolerance $\delta$. Because $t^\star$ is read from the evaluation
set, the headline saving characterises the achievable frontier rather than a
fixed deployed threshold, whereas the AUC requires no threshold and is the
selection-free measure of router quality (threshold transfer is verified in
Sec.~\ref{sec:threshtransfer}). The oracle \emph{label} threshold $\tau$ and the
deployed \emph{quality tolerance} $\delta$ are two distinct ideas, but for
simplicity we set them to the same value, $\tau=\delta=0.02$, so that the
per-document training target matches the standard used when the system is actually
deployed (Sec.~\ref{sec:oracle}).

\subsection{Cost model}
Per-document cost is token-based:
\begin{equation}
c_m(d) = \pi^{\mathrm{in}}_m\, n^{\mathrm{in}}(d) + \pi^{\mathrm{out}}_m\, n^{\mathrm{out}}_m(d),\quad m\in\{s,\ell\},
\end{equation}
where $\pi^{\mathrm{in}}_m,\pi^{\mathrm{out}}_m$ are tier $m$'s per-token
input/output prices and $n^{\mathrm{in}},n^{\mathrm{out}}_m$ the input and output
token counts. Using the published list prices of Sec.~\ref{sec:tiers}, both the
input and output price ratios between the two tiers are $5\times$, so the
per-document cost ratio is $5\times$ regardless of the input/output token mix; an
$X\%$ cost saving corresponds to avoiding $X\%$ of that $5\times$ premium. Our
second model pair (Sec.~\ref{sec:crosspair}) uses a $3\times$ ratio, so the
reported savings are not an artifact of one particular price gap.

\subsection{Pipeline}
\emph{Offline:} documents $\rightarrow$ OCR $\rightarrow$ features $\rightarrow$
run both tiers $\rightarrow$ field-level F1 $\rightarrow$ gap $g(d)$ $\rightarrow$
labels $y(d)$ $\rightarrow$ train the router. \emph{Inference:} document
$\rightarrow$ OCR $\rightarrow$ features $\rightarrow$ $p(d)$ $\rightarrow$
threshold $t$ $\rightarrow$ cheap or expensive extractor. Baselines follow the
same path with the routing decision replaced by a fixed rule.

\subsection{Pre-inference features}
\label{sec:features}
We compute 13 interpretable, document-intrinsic features
(Table~\ref{tab:features-full}), each cheap to compute and available
\emph{before} any extraction call, spanning four families: OCR quality, image
quality, layout, and content/structure. Every feature derives only from the page
image, its OCR text, or OCR box geometry, never from ground-truth
annotations, so the router is strictly pre-inference. An earlier 16-feature set
additionally used three annotation-derived signals (\texttt{label\_entropy},
\texttt{label\_diversity}, \texttt{section\_count}); we remove them to eliminate
label leakage, and find this costs almost nothing (pooled 5-fold CV AUC $0.730$
clean vs.\ $0.731$ leaky): the leakage was not load-bearing. We also
exclude a composite \texttt{complexity\_score} used in early experiments because it
is collinear with its own components, which makes feature attribution unreliable.

\begin{table}[t]\centering\small
\caption{The 13 pre-inference, document-intrinsic features, by family. All are computed before any extraction call, from the image, OCR text, or OCR box geometry only, never from ground-truth annotations.}
\label{tab:features-full}
\begin{tabular}{@{}l p{0.56\columnwidth}@{}}
\toprule
Feature & Description \\
\midrule
\multicolumn{2}{@{}l}{\textit{OCR quality}}\\
\texttt{ocr\_conf}          & Mean per-token OCR confidence \\
\texttt{ocr\_std}           & Std.\ of per-token OCR confidence \\
\texttt{ocr\_stage}         & Preprocessing stage needed to read the image \\
\texttt{short\_token\_ratio} & Fraction of tokens $\le$2 characters \\
\texttt{inv\_chars\_per\_word} & Inverse mean chars per token (fragmentation) \\
\midrule
\multicolumn{2}{@{}l}{\textit{Image quality}}\\
\texttt{blur\_score}        & Variance of the image Laplacian (sharpness) \\
\texttt{image\_contrast}    & Global pixel-intensity contrast \\
\texttt{word\_height\_cv}   & Coef.\ of variation of word-box heights \\
\midrule
\multicolumn{2}{@{}l}{\textit{Layout}}\\
\texttt{crowded\_line\_frac} & Fraction of lines with $>$3 words \\
\texttt{line\_density}      & Text lines per unit page height \\
\texttt{aspect\_ratio}      & Page height / width \\
\midrule
\multicolumn{2}{@{}l}{\textit{Content / structure}}\\
\texttt{item\_density}      & Line-items per text line (table density) \\
\texttt{tokens}             & Total OCR token count \\
\bottomrule
\end{tabular}
\end{table}

\section{Experimental Setup}
\subsection{Datasets}
\label{sec:datasets}
Table~\ref{tab:datasets} lists six datasets spanning five document genres, chosen
to span the two diagnostic axes, routing headroom and feature-detectable
difficulty, rather than a single domain. Receipts (CORD~\cite{cord},
SROIE~\cite{sroie}) and degraded political ad-buy forms
(DeepForm~\cite{deepform}) are the genres where routing works; invoices
(DocILE~\cite{docile}), photographed nutrition labels (POIE~\cite{poie}), and
registration forms (VRDU~\cite{vrdu}) are the diagnostic's negative and borderline
cases. Each genre's router is trained and evaluated \emph{within genre} on its own
train/test split (receipts pool CORD+SROIE); we never train a single router across
genres, because cross-genre transfer fails (Sec.~\ref{sec:transfer}).

\begin{table}[t]\centering\small
\caption{Six datasets across five genres, spanning the two diagnostic axes. ``Capture'' summarizes how the page image is produced (a proxy for whether difficulty is visible in observable features); ``Routes?'' is the study's outcome. Receipts and DeepForm satisfy both conditions; the rest each fail one (or are borderline).}
\label{tab:datasets}
\begin{tabular}{@{}lllrl@{}}
\toprule
Dataset & Genre & Capture & $N$ & Routes? \\
\midrule
CORD     & Receipts          & photo   & 900 & yes \\
SROIE    & Receipts          & scan    & 973 & yes \\
DeepForm & Ad-buy forms      & fax     & 800 & yes (non-receipt) \\
DocILE   & Invoices          & digital & 600 & no (no signal) \\
POIE     & Nutrition labels  & photo   & 120 & no (no headroom) \\
VRDU     & Registration      & scan    & 500 & marginal \\
\bottomrule
\end{tabular}
\end{table}

\subsection{Model tiers}
\label{sec:tiers}
We instantiate the small and large tiers with Claude Haiku~4.5 (\$1/\$5 per
MTok) and Claude Opus~4.8 (\$5/\$25 per MTok), a clean $5\times$ cost ratio. We
deliberately use a same-provider pair, which helps eliminate confounds that might
arise from the API, such as differences in prompt formatting, JSON-output
variance, or instruction-following, and it helps ensure both tiers follow
instructions in the same way. Both tiers receive an identical schema-conditioned
tool-use prompt, which keeps the comparison consistent. The oracle threshold and
routing classifier are then fit to this pair's gap distribution; moving to a new
pair requires refitting on that pair's gaps. We test this transfer-through-refitting
by running a second model pair, Haiku~4.5 vs.\ Claude Sonnet~4.6 (\$3/\$15 per
MTok, a $3\times$ ratio), in Sec.~\ref{sec:crosspair}.

We use generic, zero-shot extraction with no dataset-specific fine-tuning: each
tier sees only its target schema and the document. Absolute F1 will be lower than
task-specific fine-tuned models such as LayoutLMv3~\cite{layoutlmv3}; this is by
design, to simulate the realistic scenario where one general LLM serves many
genres without bespoke training. The distinction is critical: it makes
clear that routing is independent of absolute extractor quality and concerns only
the \emph{relative} gap between tiers.

\subsection{Oracle labeling}
\label{sec:oracle}
A document is labeled \emph{large-required} if
$\mathrm{F1}_{\text{large}} - \mathrm{F1}_{\text{small}} > \tau$, else
\emph{small-sufficient}. We set the label threshold equal to the deployment
tolerance, $\tau=\delta=0.02$: a document is then large-required exactly when
routing it to the small model would, on that document, lose more quality than the
tolerance allows. The classifier's training target is thus the per-document
analogue of the aggregate deployment constraint in Eq.~\eqref{eq:operating},
rather than an arbitrary cutoff. We report a sensitivity sweep over
$\tau \in \{0.01, 0.02, 0.05\}$ (Table~\ref{tab:tau}). F1 is multiset
(label, value); values are matched with a \emph{field-type canonical} rule
(money$\to$numeric equality, dates$\to$date-component, text$\to$normalized exact
match), which prevents currency prefixes, appended timestamps, and thousands
separators from being scored as extraction errors. If we skip this normalization,
a strict exact-match scorer calls harmless formatting changes ``wrong,'' so
correct extractions look like failures and the F1-gap becomes misleading. We
report strict exact-match F1 alongside the canonical figures for transparency.

\subsection{Classifier and baselines}
The router is a calibrated random forest (isotonic). We compare against
always-small, always-large, random-at-matched-cost, a single-feature
(OCR-confidence) rule, and an oracle upper bound.

\paragraph{Reproducibility}
All experiments after oracle labeling are local and deterministic: OCR uses
Tesseract and extraction uses pinned model snapshots
(\texttt{claude-haiku-4-5-20251001}, \texttt{claude-opus-4-8}). Extraction uses
the provider's default sampling (we do not set a temperature) with a single
response per document, cached per (tier, document). Extractions are the only paid,
stochastic step; once cached, every downstream analysis is deterministic and free
to reproduce. Because each F1 gap is a single stochastic draw,
small per-document gaps carry some decoding variance; a re-extraction pilot
measures this variance directly (per-document F1 std $\approx0.02$) and quantifies
the resulting oracle-label instability near $\tau$ (Sec.~\ref{sec:limitations}).
Routing overhead is negligible: feature extraction reuses the OCR pass the
pipeline already performs, and classifier inference is sub-millisecond, both
dominated by the extraction call. Code, the feature extractor, and the
oracle-labeling and routing scripts are released at
\url{https://github.com/sreerekha3547/complexity-aware-routing}.

\section{Results}
\label{sec:results}

We organize the evidence around three questions: does routing help (is there
headroom?), is it predictable (can pre-inference features identify the hard
documents?), and is it transferable (can a router generalize to another genre?).
Table~\ref{tab:diagnostic} summarizes these for five genres; only the two that
satisfy the first two conditions support routing (receipts, DeepForm), while the
others each miss one and none transfers to another genre. The following sections
detail each row: receipts (Sec.~\ref{sec:receipts}) as the clearest within-genre
evidence, DeepForm (Sec.~\ref{sec:deepform}) as the most compelling non-receipt
case, router simplicity (Sec.~\ref{sec:textbase}), and transfer
(Sec.~\ref{sec:transfer}).

\begin{table}[t]\centering\small
\caption{The diagnostic across five genres. \emph{Headroom} is the fraction of documents the cheap tier fails ($g>\tau$) and \emph{AUC} is the within-genre routing AUC (5-fold CV) from pre-inference features; both are computed over the full within-genre sample (train and test pooled), so they differ from the held-out, per-dataset figures reported in the subsections (e.g.\ Table~\ref{tab:oracle}). Routing helps only where \emph{both} are favorable (receipts, DeepForm). Invoices have ample headroom but are not predictable (semantic difficulty, invisible to features); nutrition labels are predictable but have almost no headroom (the cheap tier is near ceiling); VRDU is weak on both.}
\label{tab:diagnostic}
\begin{tabular}{@{}llrrl@{}}
\toprule
Genre & Capture & Headroom & AUC & Routes? \\
\midrule
Receipts  & photo/scan & 44\% & 0.71 & yes \\
DeepForm  & fax        & 41\% & 0.91 & yes \\
Invoices  & digital    & 67\% & 0.52 & no (not predictable) \\
Nutrition & photo      & 16\% & 0.63 & no (no headroom) \\
VRDU      & scan       & 21\% & 0.60 & marginal \\
\bottomrule
\end{tabular}
\end{table}

\subsection{Receipts: clearest within-genre evidence}
\label{sec:receipts}
On pooled CORD+SROIE, the 13-feature model reaches a held-out ROC-AUC of 0.707
(95\% CI $[0.656, 0.758]$, 2000 bootstrap resamples), compared with 0.625 for
logistic regression on the same features and 0.551 for OCR confidence alone
(Table~\ref{tab:metrics}). The feature set carries information a linear model does
not fully capture ($+0.082$ over logistic regression). We return in
Sec.~\ref{sec:textbase} to the fact that a simple text router matches this, which
is why we frame the features as interpretable \emph{inputs} rather than the
contribution. Under the field-type canonical scorer, OCR confidence is no longer
anti-predictive, so the gain over it ($+0.156$) is less informative than it looks.
The router is well
calibrated (expected calibration error $0.061$), as required for the
probability-threshold sweep; per-feature attributions are in
Sec.~\ref{sec:interpret}.

\begin{table}[t]\centering\small
\caption{Routing discrimination on the pooled held-out test set ($N{=}447$, $\tau{=}0.02$). The complexity model beats a linear read of the same features by $+0.082$ AUC; under the canonical scorer OCR confidence is no longer below chance.}
\label{tab:metrics}
\begin{tabular}{@{}lr@{}}
\toprule
Router & ROC-AUC \\
\midrule
Calibrated random forest (ours) & \textbf{0.707} \;[0.656, 0.758] \\
Logistic regression             & 0.625 \\
Single feature (\texttt{ocr\_conf}) & 0.551 \\
\midrule
Gain (RF $-$ logistic regression) & $\mathbf{+0.082}$ \\
Gain (RF $-$ single feature)      & $+0.156$ \\
\bottomrule
\end{tabular}
\end{table}

\paragraph{Effect of the scoring protocol.}
Table~\ref{tab:scorer} reports the headline under both scorers. The canonical rule
\emph{lowers} the AUC ($0.748\!\to\!0.707$) and the savings ($36/40\!\to\!33/31\%$):
by crediting the large model for correctly reformatted values that strict
exact-match rejected (currency prefixes, appended timestamps, thousands
separators), it narrows the measured gap. The canonical scorer therefore makes the
result more conservative rather than inflating it. It also lifts the OCR-confidence
baseline from below chance
($0.463$) to $0.551$, which is
why we headline the margin over logistic regression rather than over that baseline.

\begin{table}[t]\centering\small
\caption{Effect of the scoring protocol ($\tau{=}0.02$, pooled test). The field-type canonical scorer is conservative: it lowers both the headline AUC and the savings relative to strict exact-match, because it withdraws credit the strict scorer wrongly gave the large model for reformatting correct values.}
\label{tab:scorer}
\begin{tabular}{@{}lrrr@{}}
\toprule
Scorer & RF AUC & CORD save & SROIE save \\
\midrule
Strict exact-match          & 0.748 & 36\% & 40\% \\
Field-type canonical (ours) & 0.707 & 33\% & 31\% \\
\bottomrule
\end{tabular}
\end{table}

\subsection{A routable gap exists}
\begin{figure*}[t]\centering
\includegraphics[width=\textwidth]{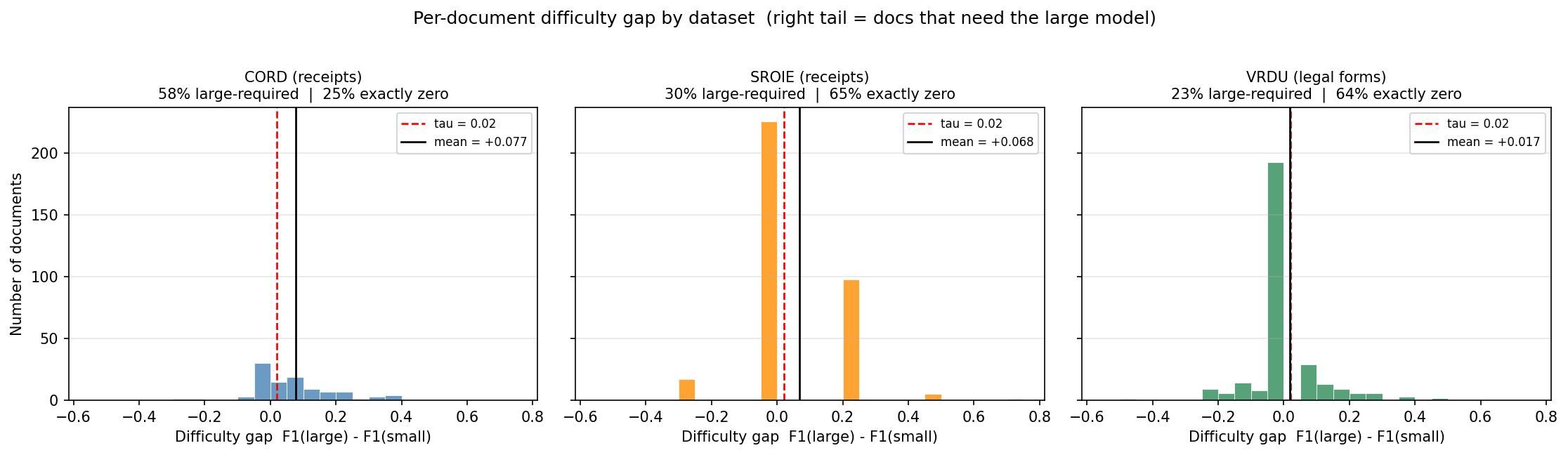}
\caption{Distribution of the per-document F1 gap (large $-$ small). CORD is right-skewed with a tail past $\tau$ (real difficulty); SROIE is bimodal, with a cluster at zero and a cluster of genuinely hard documents; VRDU is near-symmetric about zero, indicating little routing headroom.}
\label{fig:gaphist}
\end{figure*}

Table~\ref{tab:oracle} confirms that 30--58\% of receipt documents genuinely need
the large model, validating the routing premise; the per-document gap
distributions are in Fig.~\ref{fig:gaphist}. VRDU's near-zero mean gap
is itself a finding: that domain rarely needs the large model.

The gap is \emph{signed}: on 5--10\% of documents the large model scores
\emph{lower} than the small one (SROIE mean $-0.25$ on the affected $5\%$). The
canonical scorer already removes most spurious negatives: under strict
exact-match many were the large model formatting a correct value differently
(currency prefix, appended timestamp), not a genuine miss. The residual negatives
arise where the small model is already at ceiling and the large model
over-generates or departs from the target schema. Escalation is therefore not
uniformly beneficial: always-large sacrifices quality, not just cost, on these
documents, which is why we route on the signed gap and reward a router that
leaves ceiling-quality documents with the small model.

\begin{table}[t]\centering\small
\caption{Oracle gap labeling. A routable difficulty gap exists on receipts; VRDU's near-zero gap shows that domain rarely needs the large model.}
\label{tab:oracle}
\begin{tabular}{@{}lrrrr@{}}
\toprule
Dataset & F1$_{\text{small}}$ & F1$_{\text{large}}$ & Mean gap & Large-req. \\
\midrule
CORD  & 0.395 & 0.473 & $+0.077$ & 58\% \\
SROIE & 0.828 & 0.896 & $+0.068$ & 30\% \\
VRDU  & 0.790 & 0.807 & $+0.017$ & 23\% \\
\bottomrule
\end{tabular}
\end{table}

\subsection{Cost savings}
Table~\ref{tab:main} is the main result. Both CORD and SROIE clear the
predefined success criterion (within 2 F1 points of always-large while
cutting cost by $\geq$30\%). Bootstrapped $95\%$ confidence intervals on the
saving are $[26,42]\%$ (CORD) and $[27,35]\%$ (SROIE), both clear of zero; the
wider CORD interval reflects its smaller test set ($N{=}100$), so we treat the
pooled AUC as the headline discrimination metric.

\begin{table}[t]\centering\small
\caption{Cost savings at quality within 2 F1 points of always-large ($\tau{=}0.02$). The full feature model exceeds the single-feature (OCR-confidence) baseline, clearly on SROIE (31\% vs.\ 16\%) and by a smaller margin on CORD (33\% vs.\ 27\%); the discrimination gap is sharper in AUC (Table~\ref{tab:metrics}). The last column is a per-document gap-threshold \emph{policy} (route if the observed gap exceeds $\tau$), \emph{not} a true aggregate oracle; the true constrained-subset oracle is higher (CORD 61\%, SROIE 67\%), so the router operates below a genuine upper bound. VRDU and the other non-routable genres are treated in the diagnostic (Table~\ref{tab:diagnostic}) and transfer analysis (Sec.~\ref{sec:transfer}), not as savings headlines.}
\label{tab:main}
\begin{tabular}{@{}lrrr@{}}
\toprule
Dataset & Full model & Single-feature & Gap policy \\
\midrule
CORD  & \textbf{33\%} & 27\% & 32\% \\
SROIE & \textbf{31\%} & 16\% & 56\% \\
\bottomrule
\end{tabular}
\end{table}

\begin{figure}[t]\centering
\includegraphics[width=\columnwidth]{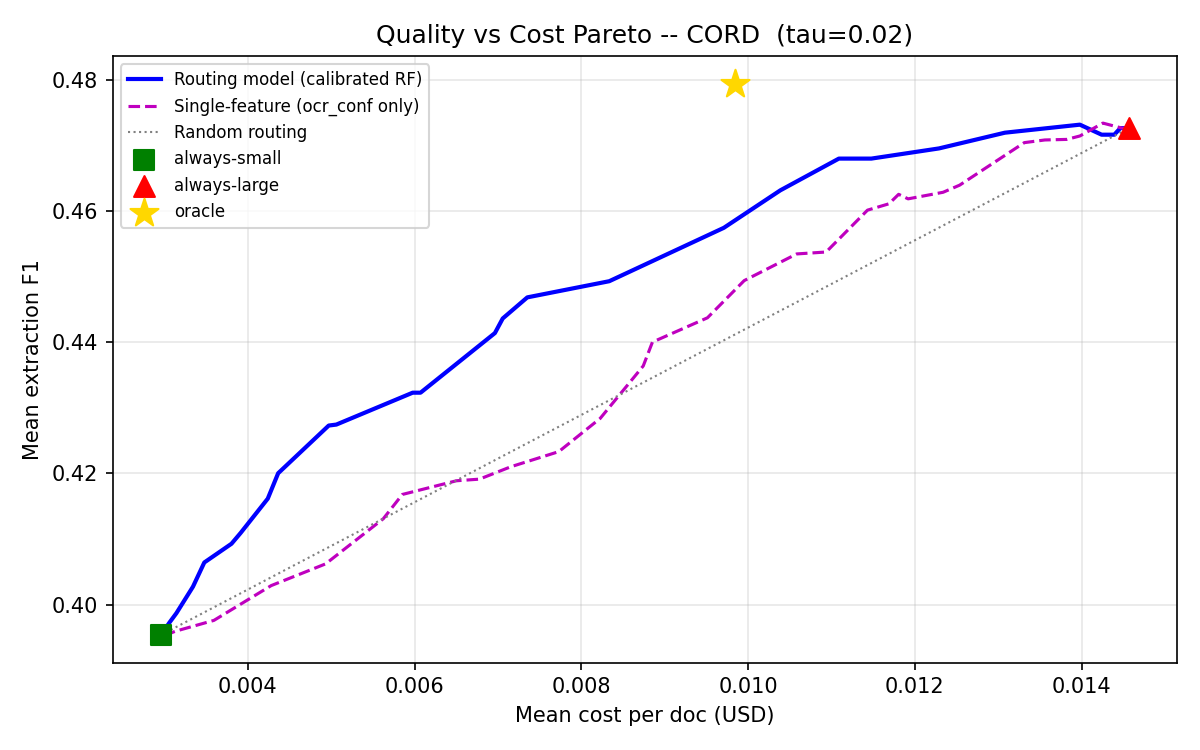}
\caption{Quality--cost Pareto frontier on CORD ($\tau{=}0.02$). The routing model sits above the single-feature and random baselines.}
\label{fig:pareto-cord}
\end{figure}

\begin{figure}[t]\centering
\includegraphics[width=\columnwidth]{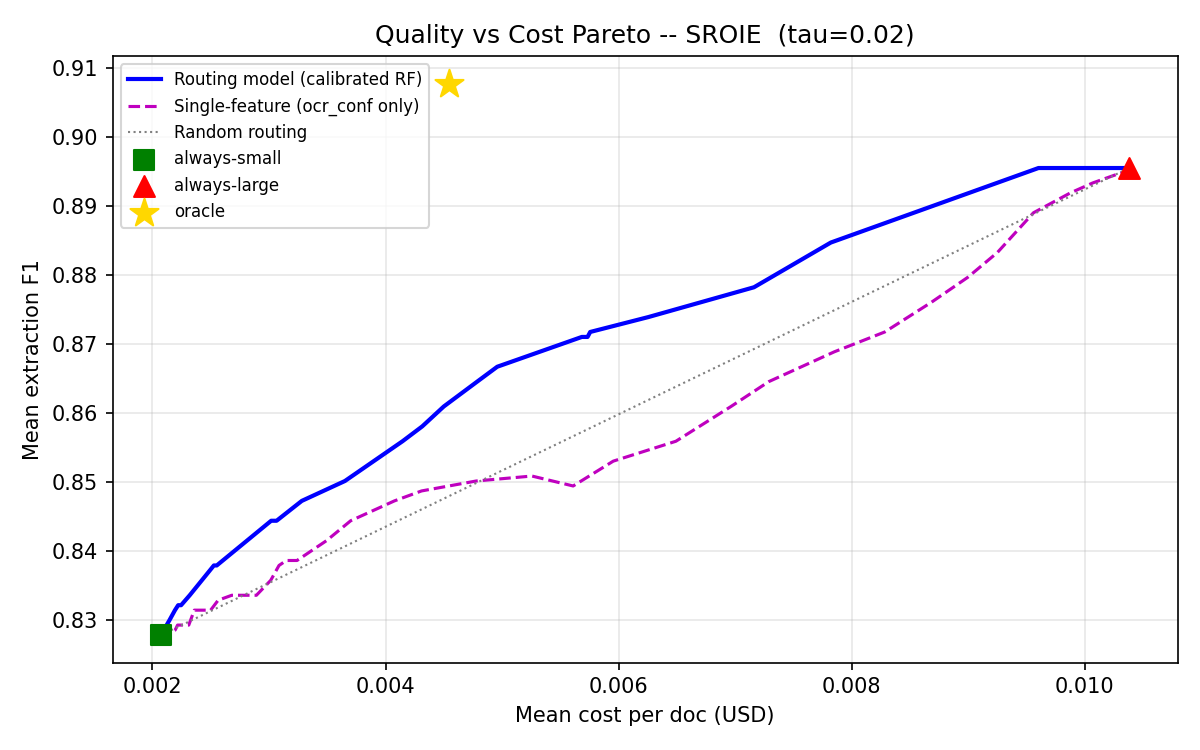}
\caption{Quality--cost Pareto frontier on SROIE ($\tau{=}0.02$).}
\label{fig:pareto-sroie}
\end{figure}

The Pareto frontiers (Fig.~\ref{fig:pareto-cord}, Fig.~\ref{fig:pareto-sroie})
show the routing model dominating both the random and single-feature baselines
across the cost range.

On CORD the full model cuts cost by 33\%, essentially matching the gap-threshold
policy at 32\%. There is no conflict here: that baseline policy decides with a
single fixed threshold per document, whereas our router works from calibrated
probabilities and can sweep thresholds, sending borderline and high-requirement
documents to the small model with only a negligible F1 loss. It can therefore
meet, and even slightly surpass, the fixed-threshold policy. This is not an
absolute upper bound; the constrained-subset oracle is higher (CORD 61\%, SROIE
67\%), and the router sits below it.

\subsection{DeepForm: routing on a non-receipt genre}
\label{sec:deepform}
Receipt difficulty is often visible in the image itself, because the documents are
photographed, dense, and frequently degraded. A stronger test is whether the same
diagnostic works on a \emph{different} genre that also satisfies both conditions.
DeepForm~\cite{deepform} is FCC political ad-buy disclosure forms: faxed,
photocopied, degraded scans of business forms, not receipts, with buried key
fields. Its capture-quality variance and
structural complexity give it feature-detectable difficulty, and its cheap-tier
failures give it headroom; the diagnostic therefore predicts it should route, and
a feature-signal-gated pilot confirmed both conditions before we ran the full
experiment.

\begin{figure}[t]\centering
\includegraphics[width=\columnwidth]{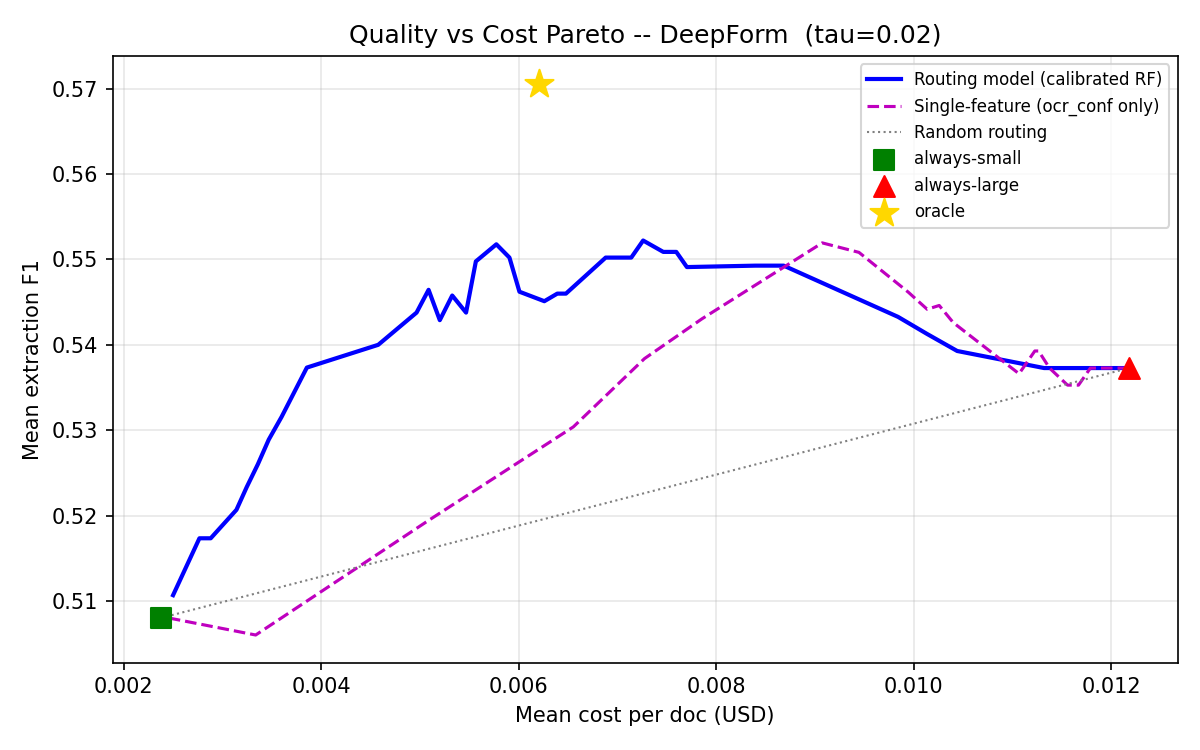}
\caption{Quality--cost Pareto frontier on DeepForm (Haiku/Opus, $\tau{=}0.02$). The router sits above the single-feature and random baselines across the cost range: a non-receipt genre where pre-inference routing works.}
\label{fig:pareto-deepform}
\end{figure}

Refit within-genre (700 train / 100 held-out test forms, five KILE fields), the
router reaches held-out AUC \textbf{0.916} (logistic regression 0.854, gain
$+0.062$), with 31\% of test documents large-required. The deployable, no-peek
policy saves \textbf{77\%} of cost at quality within the 0.02 F1 tolerance
(gap 0.020), using a threshold selected on train with a half-tolerance
generalization buffer (Table~\ref{tab:deepform}, Fig.~\ref{fig:pareto-deepform}).
The routing decisions are meaningful: on the documents the router escalates, the
large tier rescues a median $+0.20$ F1 (mean $+0.205$). Absolute F1 is low for
both tiers (small 0.62, large 0.67) because we process the first page only of
multi-page forms and $\sim$10\% of documents place a target field off page~0; this
caps absolute accuracy but not the \emph{gap}, which is what routing exploits.
Excluding the documents that are unwinnable for both tiers (those $\sim$10\%)
leaves the routing AUC essentially unchanged ($0.894$ CV on the winnable subset),
confirming the signal is genuine extraction difficulty, not a page-selection
artifact. The result holds on a second model pair (Sec.~\ref{sec:crosspair}), so
it is not specific to one expensive model.

\begin{table}[t]\centering\small
\caption{DeepForm within-genre routing (held-out test), on two model pairs. A non-receipt genre routes: high AUC, a positive RF-over-logistic margin, and a large deployable saving within the 0.02 F1 tolerance. The 3$\times$ pair saves less because the two tiers are closer on these forms (smaller gap to exploit).}
\label{tab:deepform}
\begin{tabular}{@{}lrr@{}}
\toprule
 & Haiku/Opus ($5\times$) & Haiku/Sonnet ($3\times$) \\
\midrule
Held-out AUC        & \textbf{0.916} & 0.845 \\
RF $-$ logistic     & $+0.062$ & $+0.069$ \\
Large-required      & 31\% & 22\% \\
No-peek saving      & \textbf{77\%} & 65\% \\
Quality within 0.02 & yes & yes \\
\bottomrule
\end{tabular}
\end{table}

\subsection{Comparison to a confidence cascade}
\label{sec:cascade}
A familiar alternative is a cascade: run the cheap tier first and escalate only
when its output looks unreliable (Sec.~\ref{sec:related}). That design still pays
the cheap inference on every document before deciding what to do next, whereas our
router picks a single tier up front from pre-inference features alone. Under
full-document re-extraction and the same escalation decision, pre-inference routing
is strictly cheaper than a cascade because it avoids the redundant cheap call on
escalated documents (Table~\ref{tab:cascade}); the comparison has three parts.

\emph{(i) Lower cost than a same-signal cascade.} Granting the cascade
our router's own escalation signal (its best realistic trigger) and the same
quality target, the single-tier router saves $33\%$/$39\%$ (CORD/SROIE) versus the
cascade's $21\%$/$29\%$: the cascade forfeits the saving on every escalated
document to the redundant cheap call, and can never undercut the always-small
floor.

\emph{(ii) The logprob-triggered variant is unavailable on this provider.} One
common cascade escalates on the cheap model's token logprobs; the Anthropic
Messages API exposes none, so that particular variant cannot be built here. Other
cascade triggers (schema validation, self-consistency, external verifiers) remain
possible; our cost argument applies to any of them whenever the cheap extraction
is run in full before the escalation decision.

\emph{(iii) We beat even a perfect cascade where escalation is frequent.} Against
an \emph{oracle}-triggered cascade (the best physically possible: it escalates
exactly the large-required documents, yet still pays the cheap call on every one),
the router still wins on CORD ($33\%$ vs.\ $20\%$), where escalation is frequent
($58\%$) and the redundant call is expensive. On SROIE, where escalation is rare
($30\%$), a perfect cascade would edge the router ($50\%$ vs.\ $39\%$), but that
oracle cannot be built, and the gap reflects residual headroom in the SROIE
router (well below its $56\%$ oracle route), not a cascade advantage.

\begin{table}[t]\centering\small
\caption{Cost saving vs.\ cascades ($\tau{=}0.02$). RF route is our pre-inference router; RF cascade uses the same signal but pays the cheap tier on every document; the oracle columns use the perfect (large-required) trigger.}
\label{tab:cascade}
\begin{tabular}{@{}lrrrr@{}}
\toprule
Dataset & RF route & RF cascade & Oracle cascade & Oracle route \\
\midrule
CORD  & \textbf{33\%} & 21\% & 20\% & 32\% \\
SROIE & 39\% & 29\% & 50\% & 56\% \\
\bottomrule
\end{tabular}
\end{table}

\subsection{Threshold transfer (no test peeking)}
\label{sec:threshtransfer}
The operating point in Eq.~\eqref{eq:operating} is read from the evaluation set.
To confirm the saving is not an artifact of test-set threshold selection, we
repeat the protocol without peeking: the threshold is selected on out-of-fold
\emph{training} predictions (cost-minimal subject to the quality tolerance on
train) and applied unchanged to test. On both receipt datasets it transfers within
tolerance: the blind threshold saves $33\%$ on CORD and $31\%$ on SROIE with test
quality inside the 2 F1-point tolerance (the achievable frontier is $33\%$ and
$39\%$ respectively, so the blind SROIE threshold is the more conservative of the
two). The deployable operating point is thus robust to selection on both, and the
threshold-free AUC (Table~\ref{tab:metrics}) remains our primary, selection-free
metric.

\subsection{Robustness to $\tau$}
Table~\ref{tab:tau} sweeps the oracle threshold. The AUC is steady across all
values tried ($0.707$--$0.726$); $\tau{=}0.02$ was chosen from the
quality-tolerance criterion, and the sweep shows the classifier is not
$\tau$-dependent. SROIE's large-required fraction is constant across the range
($30\%$), indicating a bimodal rather than continuous difficulty gap; its saving
rises with $\tau$ as the looser tolerance allows more aggressive routing.

\begin{table}[t]\centering\small
\caption{Sensitivity to oracle threshold $\tau$. AUC and savings are stable, confirming $\tau{=}0.02$ is not a critical design choice.}
\label{tab:tau}
\begin{tabular}{@{}lrrr@{}}
\toprule
$\tau$ & AUC (RF) & CORD saving & SROIE saving \\
\midrule
0.01 & 0.726 & 29\% & 32\% \\
0.02 & 0.707 & 33\% & 39\% \\
0.05 & 0.709 & 35\% & 40\% \\
\bottomrule
\end{tabular}
\end{table}

\subsection{Interpretability}
\label{sec:interpret}
Permutation importance for the deployed calibrated random forest (mean AUC drop on
the pooled receipt test set) ranks \texttt{inv\_chars\_per\_word} ($+0.082$),
\texttt{item\_density} ($+0.037$), and \texttt{ocr\_std} ($+0.018$) highest: token
fragmentation and content/structure density drive escalation, while the pure
image-readability features (\texttt{image\_contrast}, \texttt{blur\_score}) add
little. A per-family ablation confirms the families are complementary: training on
any single family (OCR, image, layout, or content) underperforms the full 13-feature
set ($0.707$), with layout and content the strongest and image the weakest. This
attribution is why we keep the features despite the text baseline matching them
(Sec.~\ref{sec:textbase}): they say \emph{which} document properties make
extraction hard, which is what turns routing into a diagnostic.

\subsection{The router can be simple: a text baseline}
\label{sec:textbase}
Our features are interpretable, but are they \emph{necessary}? We compare them
against a content baseline: a TF-IDF / latent-semantic (LSA) router over the same
OCR text, also pre-inference (no model call), but treating the document as a bag
of words rather than a structured artifact. Across the five genres neither router
dominates (Table~\ref{tab:textbase}): the text router edges the features on
receipts ($0.80$ vs.\ $0.71$) and DeepForm ($0.91$ vs.\ $0.89$; difference
$-0.015$, 95\% CI $[-0.033, +0.003]$), the features edge text on nutrition ($0.63$
vs.\ $0.55$), the two tie on VRDU, and both are near chance on invoices. We had
expected the image/layout features to \emph{beat} text on the capture-degraded
DeepForm forms; they do not, because degraded scans produce garbled OCR tokens
that a bag-of-words model already captures.

We draw the honest conclusion. The engineered features are not a superior routing
signal, and we do not claim they are. What the comparison shows is stronger for
the thesis: two \emph{completely different} cheap pre-inference routers (image and
layout features, and bag-of-words text) succeed on the same genres (receipts,
DeepForm) and fail on the same ones. The binding constraint is the genre's two
conditions, not the router. We report the interpretable features as the primary
router because their signals map onto the mechanism (capture quality, structure)
and thus let us \emph{state and predict} the diagnostic (which a black-box text
router cannot), not because they route better.

\begin{table}[t]\centering\small
\caption{Engineered features vs.\ a text (TF-IDF/LSA) baseline: within-genre 5-fold CV AUC. Neither cheap pre-inference router dominates across genres (text edges on receipts/DeepForm, features edge on nutrition, both near chance on invoices), evidence that the constraint is the document, not the router.}
\label{tab:textbase}
\begin{tabular}{@{}lrr@{}}
\toprule
Genre & Features (13) & Text (LSA) \\
\midrule
Receipts  & 0.71 & 0.80 \\
DeepForm  & 0.89 & 0.91 \\
Invoices  & 0.52 & 0.57 \\
Nutrition & 0.63 & 0.55 \\
VRDU      & 0.60 & 0.59 \\
\bottomrule
\end{tabular}
\end{table}

\subsection{Transfer: the router is deployment-specific}
\label{sec:transfer}
A single router that worked across corpora would be ideal; it does not exist. We
train on each genre and evaluate on every other (Table~\ref{tab:transfer}). The
within-genre diagonal is strong where routing works (receipts $0.71$, DeepForm
$0.91$) but the off-diagonal collapses to near chance (mean $0.56$). The effect is
not merely cross-genre: even two datasets of the \emph{same} genre do not transfer,
with CORD$\to$SROIE $0.54$ and SROIE$\to$CORD $0.55$, no better than cross-genre
pairs. Difficulty features mean different things across corpora, so the learned
mapping does not carry. The practical consequence is that the router is
\emph{refit per deployment} on that corpus's own labeled pilot, which is cheap (a
few dollars) and, where the conditions hold, works; the receipts result is pooled
(trained on CORD and SROIE together), i.e.\ within-distribution, not a claim of
genre-level transfer.

The failing genres reinforce the point rather than transfer. A DeepForm-trained
router scores $0.49$ on invoices: invoice difficulty is semantic and invisible to
these features, the same reason invoices' within-genre AUC is only $0.52$
(Table~\ref{tab:diagnostic}). VRDU is the borderline case: weak signal both
in-domain ($0.60$) and under transfer, with only $21\%$ headroom, and behaviorally
the router escalates $\sim$0\% of it, declining to spend rather than fabricating
savings. Neither is a routing success; both are evidence for \emph{when routing is
not worthwhile}, which is part of the contribution.

\begin{table}[t]\centering\small
\caption{Cross-genre transfer AUC (rows train, columns test; diagonal is within-genre 5-fold CV). Routing does not transfer across genres: the strong within-genre diagonal (receipts, DeepForm) collapses off-diagonal, so the router is refit per genre. Invoices and nutrition are omitted here as test targets (semantic-difficulty / small-sample off-diagonals are unreliable; see appendix); their within-genre results are in Table~\ref{tab:diagnostic}.}
\label{tab:transfer}
\begin{tabular}{@{}lrrr@{}}
\toprule
Train $\downarrow$ / Test $\rightarrow$ & Receipts & DeepForm & VRDU \\
\midrule
Receipts & \textbf{0.71} & 0.62 & 0.51 \\
DeepForm & 0.54 & \textbf{0.91} & 0.56 \\
VRDU     & 0.55 & 0.60 & \textbf{0.60} \\
\bottomrule
\end{tabular}
\end{table}

\subsection{Cross-pair generalization}
\label{sec:crosspair}
\begin{figure*}[t]\centering
\includegraphics[width=0.49\textwidth]{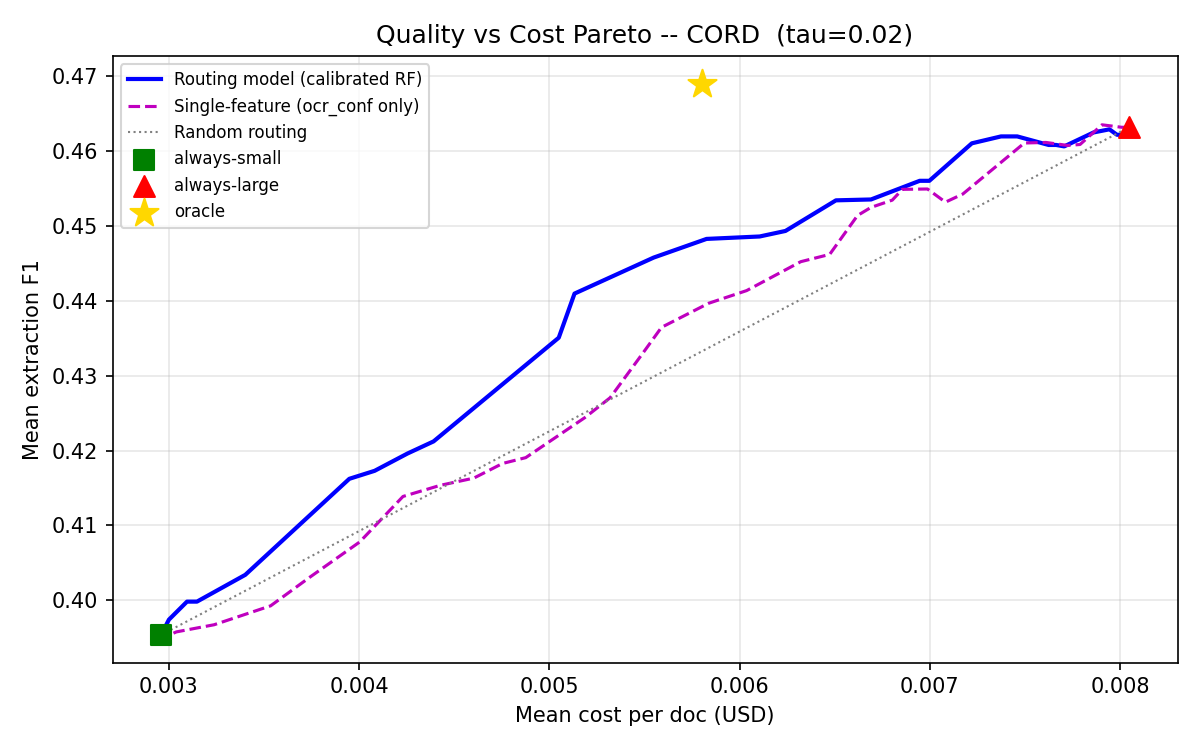}\hfill
\includegraphics[width=0.49\textwidth]{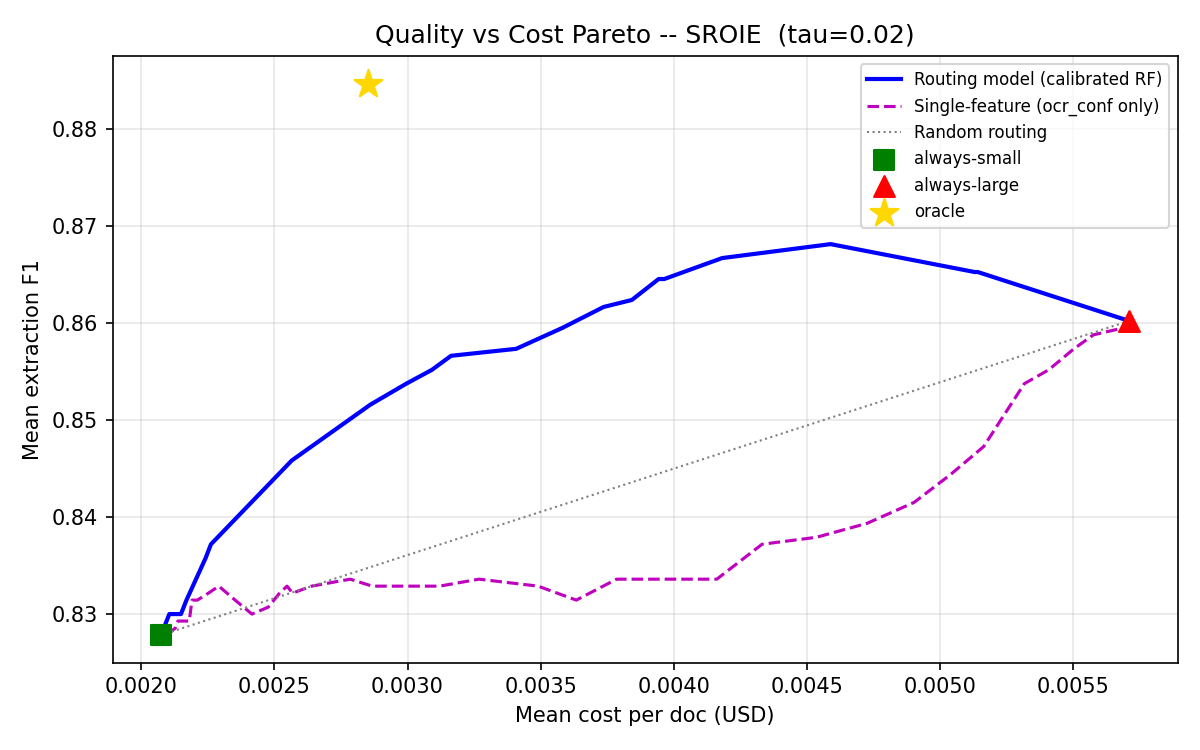}
\caption{Quality--cost Pareto frontiers for the second model pair (Haiku~4.5 vs.\ Claude Sonnet~4.6, a $3\times$ cost ratio) on CORD (left) and SROIE (right), $\tau{=}0.02$. Refit on this pair's gaps, the router again dominates the single-feature and random baselines across the cost range: cross-pair generalization holds at a different cost ratio.}
\label{fig:pareto-pair2}
\end{figure*}

The reusability claim (refit the router per model pair) is only meaningful if it
transfers. We test a second pair, Haiku~4.5 vs.\ Claude Sonnet~4.6, at a $3\times$
cost ratio (versus $5\times$ for Haiku/Opus). Sonnet is closer to Haiku than Opus
is, so the difficulty gap is smaller (mean $+0.068$ on CORD, $+0.032$ on SROIE),
yet a routable gap remains. Refitting the \emph{same} 13 features on this pair's
gaps (Table~\ref{tab:crosspair}, Fig.~\ref{fig:pareto-pair2}), the router reaches
pooled AUC $0.753$, higher than the primary pair, and saves $22\%$ (CORD) and
$51\%$ (SROIE). The larger SROIE saving follows from the smaller gap: only $22\%$
of SROIE receipts need Sonnet, so most route to cheap Haiku within tolerance.
DeepForm shows the same pair-robustness (Table~\ref{tab:deepform}): held-out AUC
$0.845$ under Haiku/Sonnet versus $0.916$ under Haiku/Opus, with the
RF-over-logistic margin holding at $+0.069$.

The margin over the linear baseline is stable across pairs and cost ratios: RF
exceeds logistic regression by $+0.082$/$+0.079$ on receipts and $+0.062$/$+0.069$
on DeepForm. What is reusable is the \emph{routing signal and the diagnostic}, not
a fixed threshold or a single model pair: the threshold is refit per pair (and per
genre), and doing so works.

\begin{table}[t]\centering\small
\caption{Cross-pair generalization on receipts. The same 13 features, router refit on each pair's gaps. Savings are the frontier values at $\tau{=}0.02$ (the deployable no-peek headline in Table~\ref{tab:main} is more conservative). The RF-over-logistic-regression margin is stable across the $5\times$ and $3\times$ cost ratios; DeepForm shows the same pattern (Table~\ref{tab:deepform}).}
\label{tab:crosspair}
\begin{tabular}{@{}llrrrr@{}}
\toprule
Model pair & Ratio & AUC & RF$-$LR & CORD & SROIE \\
\midrule
Haiku / Opus~4.8   & $5\times$ & 0.707 & $+0.082$ & 33\% & 39\% \\
Haiku / Sonnet~4.6 & $3\times$ & 0.753 & $+0.079$ & 22\% & 51\% \\
\bottomrule
\end{tabular}
\end{table}

\section{Limitations}
\label{sec:limitations}
The contribution is the diagnostic (measuring headroom, testing feature
predictability, and evaluating transfer), not a claim that any particular router is
optimal. A bag-of-words text router matches our engineered features on every genre
(Sec.~\ref{sec:textbase}); we prefer the interpretable features for their
explanatory value, but they are not a superior routing signal, and better routers
(including feature/text fusion) are open. Our positive evidence is concentrated in
two genres: receipts (CORD, SROIE) and degraded ad-buy forms (DeepForm); these
span photographed receipts and faxed business forms, but broader coverage of the
\emph{positive} case is valuable future work. Routing does not transfer across
genres (Sec.~\ref{sec:transfer}): a router must be refit per genre, which the
diagnostic pilot makes cheap but which precludes a single universal router. On
DeepForm we process the first page only of multi-page forms, which caps absolute F1
(some target fields fall on later pages) though not the routable \emph{gap} that
routing exploits; full multi-page handling is future work. The deployed threshold
is refit per model pair and per genre; we validate the pair-refit on a second pair
(Sec.~\ref{sec:crosspair}), but broader model-pair and cross-provider coverage
remains open. We pin exact model versions
(\texttt{claude-haiku-4-5-20251001}, \texttt{claude-opus-4-8}); provider updates
may shift absolute numbers. OCR confidence is from Tesseract and is not calibrated
across engines, so OCR-derived features are engine-specific. By ``document'' we
mean the page artifact (its layout, image quality, and structure) as opposed to
a text query; the routing signal is a property of that artifact. Extraction was run once per document at the
provider's default temperature rather than averaged over samples. A re-extraction
pilot measures the resulting decoding variance directly (per-document F1 std
$\approx0.02$) and the oracle-label instability it induces near the boundary:
about $20\%$ of documents within $0.10$ F1 of $\tau$ flip label under $k$-draw
averaging, almost all zero-gap documents that carry a small positive expected gap.
Because $\tau{=}0.02$ sits within this decoding-noise floor, the binary label is
inherently soft near the boundary; the threshold-free AUC is nonetheless stable
across $\tau$ (Table~\ref{tab:tau}), so the primary metric does not depend on where
the boundary falls. We do not evaluate on
FUNSD~\cite{funsd} despite its prominence in document AI: its annotations are
structural entity-linking labels (header/question/answer/other) rather than
named key--value fields, so there is no per-field value to score with our
extraction-F1 protocol, making it incompatible with the routing task as posed.

\section{Conclusion}
We recast pre-inference document routing as a question of \emph{when} it works and
give a diagnostic (routable headroom and feature-detectable difficulty) that a
small labeled pilot can answer before committing at scale. Evaluated across five
genres, the diagnostic separates the genres where routing helps (receipts and
degraded ad-buy forms, a non-receipt genre, where a calibrated router cuts cost by
$31$--$33\%$ and $77\%$ at quality within $0.02$ F1 of always-large) from those
where it does not (clean digital invoices with no feature-detectable signal,
near-ceiling nutrition labels with no headroom, and low-headroom registration
forms). The routing signal is available in cheap pre-inference representations,
both interpretable document features and OCR text; within the engineered model,
token fragmentation and layout density matter most. The router is
deployment-specific, refit per corpus, and does not transfer across datasets even
within a genre. Under full-document re-extraction, pre-inference routing is strictly
cheaper than a cascade, and beats even a perfect-trigger cascade where escalation
is frequent, because a cascade pays the cheap-model call on every document; the
logprob-triggered variant is unavailable on this provider. The results hold across
two model pairs at $5\times$ and $3\times$ cost ratios. The practical takeaway is a
checklist: before deploying routing on a new corpus, measure headroom and feature
predictability on a small labeled pilot, and if either is absent, do not route.
Extending the diagnostic across providers and to vision-language extractors is the
natural next step.


\end{document}